# Recognizability of Individual Creative Style Within and Across Domains: Preliminary Studies

**Liane Gabora (liane.gabora@ubc.ca)**
Department of Psychology, University of British Columbia
Okanagan campus, 3333 University Way
Kelowna BC, V1V 1V7, CANADA

## Abstract

It is hypothesized that creativity arises from the self-mending capacity of an internal model of the world, or worldview. The uniquely honed worldview of a creative individual results in a distinctive style that is recognizable within and across domains. It is further hypothesized that creativity is domain-general in the sense that there exist multiple avenues by which the distinctiveness of one's worldview can be expressed. These hypotheses were tested using art students and creative writing students. Art students guessed significantly above chance both which painting was done by which of five famous artists, and which artwork was done by which of their peers. Similarly, creative writing students guessed significantly above chance both which passage was written by which of five famous writers, and which passage was written by which of their peers. These findings support the hypothesis that creative style is recognizable. Moreover, creative writing students guessed significantly above chance which of their peers produced particular works of art, supporting the hypothesis that creative style is recognizable not just within but across domains.

## Introduction

The therapeutic nature of the creative process is well known. Eminent creators and laypeople alike often claim that through engagement in creative activities they gain a clearer sense of themselves as unique individuals. By making artistic choices, and observing how these choices guide subsequent thoughts about the work, eventually culminating in original, creative form, they acquire self-knowledge, and often, are left with a sense of completeness. The transformation that occurs on canvas or on the written page is said to be mirrored by a sense of personal transformation and self-discovery that occurs within.

Artists often find a style that feels as if it is 'theirs' only after periods of exploration with different media and established styles and art forms. Similarly, writers speak of transitioning from a stage in which they were merely imitating the styles of authors they admired to a stage in which they discovered their own authentic 'voice'. This sense of self-discovery may seem to the creator as real as anything he or she has ever experienced, and the transition from merely imitating others to finding one's own identifiable style is often evident to anyone exposed to an individual's creative works. But although the phenomenon of recognizable creative style seems obvious to artists themselves, and to those who appreciate what they do, it is not predicted by well-known theories of creativity.

This paper presents the results of preliminary experiments designed to test the hypothesis that creative individuals possess a distinctly recognizable creative style, and that this creative style is recognizable not just within a domain but across domains. We begin by discussing by well-known theories of creativity, and how the phenomena of individual style and 'voice' are not predicted by them. Three studies are then presented. The first two studies test the hypothesis that the phenomenon of creative style is real; that is, that creative individuals such as artists and writers genuinely exhibit a creative style that others come to associate with them. The third study tests the hypothesis that an individual's creative style is recognizable not just in one domain, but across different domains. Finally, we discuss how the findings are compatible with a new theory of creativity.

## Theories of Creativity

This section very briefly summarizes some leading theories of how the creative process works, and then presents a new theory of creativity referred to as honing theory.

### Creativity as Heuristic Search

Inspired by the metaphor of the mind as a computer (or computer program), it was proposed that creativity involves a process of heuristic search, in which rules of thumb guide the inspection of different states within a particular state space (set of possible solutions) until a satisfactory solution is found (Eysenck, 1993; Newell, Shaw & Simon 1957; Newell & Simon 1972). In heuristic search, the relevant variables are defined up front; thus the state space is generally fixed. Examples of heuristics include breaking the problem into sub-problems, hill-climbing (reiteratively modifying the current state to look more like the goal state), and working backward from the goal state to the initial state. A variation on this is the view that creativity involves heuristics that guide the search for, not a possibility within a given state space, but a new state space itself (*e.g.,* Boden, 1990; Kaplan & Simon, 1990, Ohlsson, 1992). That is, it involves switching from one representation of the problem to another, sometimes referred to as *restructuring* (Weisberg, 1995).

### The Expertise View of Creativity

Some posit that creativity involves everyday thought processes such as remembering, planning, reasoning, and restructuring; no special or unconscious thought processes need be postulated (Perkins, 1981; Weisberg, 2006). This is sometimes referred to as the *expertise view* of creativity because it stresses the extent to which creative acts draw upon familiarity with a particular domain of knowledge. Thus this view in particular is associated with the notion that creativity is highly domain-specific; expertise in one domain is not viewed as enhancing creativity in another domain. The expertise view is also associated with the notion that the creative process result in products that are largely derivative, or *reproductive* (as Weisberg puts it), as opposed to genuinely new, or *productive*.

### The Darwinian Theory of Creativity

Another approach to modeling the creative process involves framing it in Darwinian terms. While some philosophers describe the growth of knowledge as Darwinian merely in the sense that conjectures must be refutable, *i.e.*, able to be selected against (Popper, 1963; Lorenz, 1971), Campbell (1960) goes further, arguing that a stream of creative thought is a Darwinian process. The basic idea is that we generate new ideas through 'blind' variation and selective retention (abbreviated BVSR): 'mutate' the current thought a multitude of different ways, select the fittest variant(s), and repeat this process until a satisfactory idea results. The variants are 'blind' in the sense that the creator has no subjective certainty about whether they are a step in the direction of the final creative product.

Currently the Darwinian view of creativity is most closely associated with Simonton (1998, 1999a,b, 2007a,b), who views creativity as essentially a trial-and-error process in which the most promising 'blindly' generated ideational variants are selected for development into a finished product. It should be noted that the endeavor to apply natural selection to creative thought is not without critics (Dasgupta, 2004; Eysenck, 1995; Gabora, 2005, 2007; Sternberg, 1998, Thagard, 1980; Weisberg, 2000, Weisberg & Haas, 2007). Nevertheless, the development of a creative idea can be said to be evolutionary in the very general sense that it exhibits descent with adaptive modification.

### The Honing Theory of Creativity

Central to the *honing theory* of creativity is the notion of a *worldview,* by which we mean one's internal model of the world, as well as one's values, attitudes, predispositions, and habitual patterns of response (Gabora, 2000, 2004, 2008, Gabora & Aerts, 2009). Honing theory posits that creativity arises due to the *self-organizing, self-mending* nature of a worldview, and that it is by way of the creative process the individual hones (and re-hones) an integrated worldview. Honing theory places equal emphasis on the externally visible creative outcome and the internal cognitive restructuring brought about by the creative process. Indeed one factor that distinguishes it from other theories of creativity is that it focuses on not just restructuring as it pertains to the conception of the task, but as it pertains to the worldview as a whole. When faced with a creatively demanding task, there is an interaction between the conception of the task and the worldview. The conception of the task changes through interaction with the worldview, and the worldview changes through interaction with the task. This interaction is reiterated until the task is complete, at which point not only is the task conceived of differently, but the worldview is subtly or drastically transformed. Thus one distinguishing feature of honing theory is that the creative process reflects the natural tendency of a worldview to seek integration or consistency amongst both its pre-existing and newly-added components, whether they be ideas, attitudes, or bits of knowledge; it mends itself as does a body when injured.

## The Recognizability of Creative Style

Theories of creativity based on heuristic search, the acquisition of expertise, or chance, random processes, such as BVSR, give no reason to expect that the act of creation leads to a clearer or more integrated sense of self, or that the works of a particular creator should exhibit a unique and recognizable style. This is particularly so if, as is often claimed, creativity is strongly domain-specific (Baer, 1998; Sawyer, 2006; Weisberg, 2006). If creativity is limited to a particular domain then why should it result in a global sense of wellbeing or integration?

Claims about the domain-specificity of creativity are based largely on findings that correlations amongst alternative measures of creativity are small, and expertise or eminence with respect to one creative endeavor is rarely associated with expertise or eminence with respect to another (e.g. Getzels & Jackson, 1962). Thus, for example, creative scientists rarely become famous artists or dancers. The focus of these studies is squarely on expertise or eminence as evidence of creative achievement. But what if creative achievement is measured not by expertise or eminence but by having found a way to express *what is genuine and unique about us* through whatever media we have at a given time at our disposal? One might expect that an artist's or scientist's personal style comes through in how he or she prepares a meal or decorates a room, what creativity researchers refer to as little-c (Richards, Kinney, Benet, & Merzel, 1988) or mini-c (Beghetto & Kaufman, 2007) creative activities. Findings of domain-specificity in creativity may have more to do with the fact that we focus on creative achievement at a level that takes a decade or more to obtain (Simonton, 2007), as opposed to looking for evidence that creative potential and personal style transcends particular domains. In other words, looking for evidence of exceptional creativity in multiple domains is not the only or necessarily even the best way to address the question of whether creativity is domain-specific. Another way is to look for evidence that an individual exhibits a creative style in one domain that also 'comes through' when engaged in creative activities in other domains.

Although the phenomenon of recognizable style or voice is not predicted by the view that creativity is a matter of heuristic search, expertise, or Darwinian selection, it is predicted by the honing theory of creativity. We have seen that, according to honing theory, creativity is the process by which one hones a worldview, and each idea the creator comes up with is a different expression of the same underlying core network of understandings, beliefs, and attitudes. A worldview has a characteristic structure, and the creator's various outputs are reflections of that structure, and they are related to one another, and potentially pave the way for one another. Thus honing theory predicts that creative individuals have a recognizable style.

There is evidence that human creativity is more consistent with honing theory than with competing theories of creativity with respect to developmental antecedents of creativity, personality traits of creative individuals, and studies of lifespan creativity (Gabora, under revision). This paper reports on the results of creative style experiments that provide further support for the theory. The goal of the first two studies was to find empirical evidence for the common belief that there really *is* such a thing as recognizable style or voice. Although artists have no doubt this is true, it has not been studied by psychologists, and as we have seen, most theories of creativity do not predict it. The goal of the third study was to test a more controversial prediction of honing theory, the prediction that the structure of a worldview manifests in a unique and recognizable way, to varying degrees, through *different* creative outlets. Thus for example, you might recognize someone's art by knowing how they dress or decorate.

## Study 1: Within-domain Recognizability of Artistic Style

The first study tested the hypothesis that individuals who are highly familiar with the art of a given artist will recognize other works by that artist that they have not encountered before.

### Method

***Participants*** The research was conducted with 10 University of British Columbia undergraduates majoring in art who were highly familiar with five well-known artists, and with each other's art.

***Materials and Procedures*** Prior to the study, participants were instructed to bring from home a recently completed painting that they had never discussed with or shown to any of their classmates. They were asked to hide their signatures or any other identifying feature of the painting. Before the study, the paintings were examined to ensure that signatures and any other identifying features had been covered.

At the beginning of the study, the art students were shown three well-known paintings by each of five well-known artists as a refresher. The well-known artists were Picasso, Monet, Van Gogh, Dahli, and Andy Warhol. These artists were decided upon because previous discussion with the class indicated that all students were highly familiar with them. The students were then shown ten unfamiliar (rare or newly completed) works that they had not studied in class. Signatures on all artworks were covered by black tape. The art students were given a questionnaire and asked to guess which famous artist did each painting. For each answer they were also asked to state how certain they were on a 3-point scale that they had not encountered the work before.

They were also shown the paintings by their fellow classmates that they had never seen before. The rationale for showing classmates' paintings was to control for the possibility that with the well-known artists, a participant who, though not recognizing the creative voice, might guess above chance levels to which era or country the artist belonged. The only sufficiently large number of artists from the same era and locale that the students were familiar with were their own classmates.  As with the famous artists, they were asked to guess which classmate did each painting, and to state how certain they were on a 3-point scale that they had not encountered the work before.

The participants were debriefed, and the results were analyzed. If a participant had encountered a work before, or was uncertain about having encountered it before, the score for this question was not included in the analysis. Less than 5% of scores were not included in the analysis.

***Analysis*** The data were analyzed to determine if the participants correctly identified the artists at above-chance levels. First, a proportion correct score for each participant was computed. For example, if a participant correctly identified seven out of 10 possible artists, the proportion correct score for that person was .70. Then, the proportion correct score that would have been obtained on the basis on random guesses for each question was computed. For example, for the well-known artists, since there were 5 of them, the proportion correct based on random guesses was .20. One-sample t-tests were then computed comparing the average proportion correct scores to the proportion correct values that would have been obtained had participants been randomly guessing. A one-sample randomization test (Manly, 2007) was used to compute the p-levels for these t-test values, given the small sample sizes, and .05 was used as the criterion for statistical significance.

### Results

The results are divided into two sections: recognition of famous artists, and recognition of classmates' art.

***Recognition of Famous Artists*** For the task in which art students were asked which famous artist painted each painting, the mean proportion correct was .78 (SD = .12). The proportion correct that would have been obtained on the basis of random guesses was .20. This difference is statistically significant, $t(9) = 15.3$, $p < .0001$, $r$ (effect size) = .98. Thus art students were able to distinguish above chance which famous artists created pieces of art they had not seen before.

***Recognition of Classmates' Art*** A similar result was obtained with works of art by the students themselves. The mean proportion correct was .74 (SD = .29). The proportion correct that would have been obtained on the basis of random guesses is .11. This difference is significant, $t(9) = 6.8$, $p < 0.0001$, r = .92. Thus art students also correctly identified their classmates' art above chance.

## Study 2: Within-domain Recognizability of the Notion of a Writer's 'Voice'

This study tested the hypothesis that individuals who are highly familiar with the work of a given writer will recognize other works by that writer that they have not encountered before.

### Method

***Participants*** The research was conducted with seven University of British Columbia advanced undergraduate creative writing students who were highly familiar with five well-known writers, and with each other's writing.

***Materials and Procedures*** The analogous procedure to that described above for art students was used for creative writing students. Prior to the study, they had been asked to write a passage about a kitchen and a poem about a month of the year. They were explicitly asked to include no immediately identifying content in their writing (*e.g.,* no mention of surfing if it is known that they like surfing). These constituted their two pieces of writing. At the beginning of the study they were given three well-known written passages by each of ten well-known writers as a refresher. The well-known writers were Ernest Hemingway, Douglas Coupland, Emily Dickinson, Walt Whitman, Allen Ginsburg, Jack Kerouac, TS Eliot, Jane Austin, George Orwell, and Franz Kafka. These writers were chosen because previous discussion with the class indicated that all students were highly familiar with them. A sample of one of the written passages by well-known writers (in this case, Ernest Hemingway) that were provided to creative writing students as a refresher is provided in Table 1.

Table 1: Sample of written passage by well-known writer provided to creative writing students as a refresher.

"If the book is good, it is about something that you know, and is truly written, and reading it over you see that this is so, you can let the boys yip and the noise will have that pleasant sound coyotes make on a very cold night when they are out in the snow and you are in your own cabin that you have built or paid for with your work."

The creative writing students were then shown twenty rare passages that they had not studied in class. A sample of one of the passages by well-known writers (in this case, Jane Austin) is provided in Table 2.

Table 2: Sample of written passage by well-known writer provided to creative writing students as a test of their ability to recognize writer's style.

"However, here they are, safe and well, just like their own nice selves, Fanny looking as neat and white this morning as possible, and dear Charles all affectionate, placid, quiet, cheerful, good humour. They are both looking very well, but poor little Cassy is grown extremely thin, and looks poorly. I hope a week's country air and exercise may do her good. I am sorry to say it can be but a week. The baby does not appear so large in proportion as she was, nor quite so pretty, but I have seen very little of her. Cassy was too tired and bewildered just at first to seem to know anybody. We met them in the hall -- the women and girl part of us -- but before we reached the library she kissed me very affectionately, and has since seemed to recollect me in the same way."

The creative writing students were also given the two pieces of writing by each of their fellow classmates (the passage about a kitchen and the poem about a month of the year) that they had never seen before. They were given a questionnaire, and asked to guess which famous writer wrote each passage in the first set of passages, and which classmate wrote each passage in the second set. For each answer, they were also asked to state on a 3-point scale how certain they were that they had not encountered the work before.

Participants were debriefed, and the results were analyzed. As in the first study, if the participant had encountered the work before, or was uncertain about having encountered it before, the score for this question was not included in the analysis. Once again, less than 5% of scores were not included in the analysis.

### Results

The results are divided into two sections: recognition of famous writers, and recognition of classmates' writing.

***Recognition of Famous Writers*** For creative writing students exposed to passages by famous writers, the mean proportion correct was .34, (SD = .28). The proportion correct that would have been obtained on the basis of random guesses is .10. This difference is significant, $t(7) = 7.0$, $p < 0.0001$, r = .94. Thus creative writing students correctly identified above chance passages by famous writers that they had not encountered before.

***Recognition of Classmates' Writing*** A similar but less pronounced result was obtained with passages written by the students themselves. The mean proportion correct was .27 (SD = .16). The proportion correct that would have been obtained on the basis of random guesses is .14. This difference is significant, $t(7) = 2.3$, $p < 0.05$, r = .66. Thus, creative writing students also correctly identified above chance passages written by classmates.

## Study 3: Cross-domain Recognizability of Style

This experiment tested the hypothesis that familiarity with an individual's creative work in one domain facilitates recognition of that individual's creative work in another.

## Method

***Participants*** The same seven University of British Columbia advanced undergraduate creative writing students who participated in Study 2 also participated in Study 3. They were highly familiar with each other's writing, but unfamiliar with each other's art.

***Materials*** Each creative writing student brought one piece of covered art to the professor's office. They were asked to hide their signature and any other identifying feature. Before the study, the paintings were examined to ensure that signatures and other identifying features had been hidden.

***Procedure*** The students were shown unsigned art done by classmates. They were given a questionnaire and asked to guess which classmate did which piece of art. As above, for each answer they were also asked to state on a scale of 1-3 how certain they were that they had not encountered the work before. If they had seen the piece before, or thought they might have seen it before, their answer was not included in the analysis. Less than 5% of scores were discarded from the analysis.

## Results

The mean proportion correct was .39 (SD = .27). The proportion correct that would have been obtained on the basis of random guesses is .17. This difference is significant, $t(6) = 2.2$, $p < 0.03$, $r = .67$. Thus creative writing students were able to identify above chance which of their classmates created a given work in a domain *other* than writing, specifically art.

# Discussion and Conclusions

The experiments with artists and writers reported here provide support for the hypothesis that different works by the same creator exhibit a recognizable style or 'voice', and that this recognizable quality even comes through in different creative outlets. Art students were able to distinguish significantly above chance which famous artists created pieces of art they had not seen before. They also correctly identified their classmates' art significantly above chance. Similarly, creative writing students correctly identified significantly above chance passages by famous writers that they had not encountered before, and correctly identified their classmates' writing significantly above chance. Creative writing students additionally correctly identified significantly above chance works of art produced by classmates. (The opposite study, determining whether art students correctly identify written passages generated by their classmates, has not yet been carried out.)

The higher recognizability of artistic style (study 1) than writer's style (study 2) comes as a surprise. It cannot be entirely due to the famous artists coming from a wider range of eras and locales than the famous writers, for if that were the correct explanation, the recognizability of classmates' art in Study 1 and classmates' writing in Study 2 should have been comparable. Perhaps there are fewer constraints on artists, *i.e*. fewer demands that the work 'make sense', and it need not exhibit plot structure or character development. Thus there may be more acceptable ways of 'doing one's own thing'. The analysis takes into account that there were twice as many writers to choose from as artists, but in future studies the number of famous artists and writers will be the same, in order to make the studies as comparable as possible.

The results support the hypothesis that creators have a recognizable style. These findings are not predicted by theories of creativity that emphasize chance processes or the accumulation of expertise. If creative output is a matter of chance or the acquisition of expertise, then what is the source of this identifiable personal style? These findings are, however, predicted by honing theory, according to which personal style reflects the uniquely honed structure of an individual's worldview. The finding that creative writing students were able to identify above chance which of their classmates created a given work in a domain *other* than writing, specifically art, supports the prediction that creators hone a uniquely structured worldview that exhibits a style that is recognizable not just within a domain but across domains. Further experiments are underway to replicate these findings with larger groups of participants, and adapt the general procedure to study the recognizability of style within and across domains using trained jazz musicians.

It is worth pointing out explicitly how this approach, and in particular the investigation of recognizable cross-domain style, differs from typical attempts to determine to what extent higher cognition is domain-general versus domain-specific, or modular. As mentioned in the introduction, it is commonly assumed that this issue can be resolved by determining to what extent ratings of expertise in one domain are correlated with ratings of expertise in another. An unspoken assumption here is that measurements of expertise are all that is needed to detect any sort of quality that might characterize or unify an individual's creative or intellectual ventures, and indeed that the outputs of higher cognitive processes are objectively comparable. But the reality is that while manifestations of higher cognition are *sometimes* comparable, even quantitatively, often there is little objective basis for comparison. The present results suggest that higher cognition is domain general not in the sense that expertise in one enterprise guarantees expertise in another, but in the sense that there are multiple interacting venues for creative exploration and self-expression open to any individual, and through which that individual's worldview may be gleaned. It may be that our potential for cross-domain learning is only just beginning to be exploited, through ventures such as the Learning through the Arts program in Canada, in which students, for example, learn mathematics through dance, or learn about food chains through the creation of visual art. It seems reasonable that if knowledge is *presented* in compartmentalized chunks, students end up with a compartmentalized understanding of the world, while if knowledge were presented more holistically, a more integrated kind understanding may be possible.

## Acknowledgments

This work is funded by grants from the Social Sciences and Humanities Research Council of Canada (SSHRC) and the GOA program of the Free University of Brussels. The author wishes to thank Brian O'Connor for insightful discussion and statistical help.